\documentclass[11pt,a4paper]{article}
\usepackage{authblk}
\usepackage{times}
\usepackage{latexsym}
\usepackage{epsfig}
\usepackage{graphicx}
\usepackage{amsmath}
\usepackage{amssymb}
\usepackage{multirow}
\usepackage{caption}
\usepackage{subcaption}
\usepackage{soul}
\usepackage{url}
\usepackage[autostyle, english = american]{csquotes}
\usepackage[hyperref]{acl2019}
\usepackage{footnote}

\aclfinalcopy 
\def\aclpaperid{132} 

\newcommand{\etal}{\textit{et al}.}
\newcommand{\ie}{\textit{i}.\textit{e}.}
\newcommand{\eg}{\textit{e}.\textit{g}.}
\title{Weakly-Supervised Spatio-Temporally  Grounding Natural Sentence in Video}

\author[Chen \etal]{
Zhenfang Chen$^1$\thanks{\quad Work done while Zhenfang Chen was a Research Intern with Tencent AI Lab.}\quad
Lin Ma$^2$\thanks{\quad Corresponding authors.}\quad Wenhan Luo$^{2\dagger}$\quad Kwan-Yee K. Wong$^1$ \\
$^1$The University of Hong Kong  \quad $^2$Tencent AI Lab  \\
\texttt{\{zfchen, kykwong\}@cs.hku.hk} \\
\texttt{\{forest.linma, whluo.china\}@gmail.com}
  
}

\begin{document}

\maketitle

\begin{abstract}
In this paper, we address a novel task, namely \textit{weakly-supervised spatio-temporally grounding natural sentence in video}. Specifically, given a natural sentence and a video, we localize a spatio-temporal tube in the video that semantically corresponds to the given sentence, with no reliance on any spatio-temporal annotations during training. First, a set of spatio-temporal tubes, referred to as instances, are extracted from the video. We then encode these instances and the sentence using our proposed attentive interactor which can exploit their fine-grained relationships to characterize their matching behaviors. Besides a ranking loss, a novel diversity loss is introduced to train the proposed attentive interactor to strengthen the matching behaviors of reliable instance-sentence pairs and penalize the unreliable ones. Moreover, we also contribute a dataset, called VID-sentence, based on the ImageNet video object detection dataset, to serve as a benchmark for our task. Extensive experimental results demonstrate the superiority of our model over the baseline approaches.
Our code and the constructed VID-sentence dataset are available at:~\texttt{~\url{https://github.com/JeffCHEN2017/WSSTG.git}}.
\end{abstract}

\section{Introduction}
Given an image/video and a language query, image/video grounding aims to localize a spatial region in the image~\cite{plummer2015flickr30k,yu2017joint,yu2018mattnet} or a specific frame in the video~\cite{zhou2018weakly} which semantically corresponds to the language query. Grounding has broad applications, such as text based image retrieval~\cite{chen2017amc,ma2015multimodal}, description generation~\cite{wang2018reconstruction,rohrbach2017generating,wang2018bidirectional}, and question answer~\cite{gao2018motion,ma2016learning}. Recently, promising progress has been made in image grounding~\cite{yu2018mattnet,chen2018realtime,zhang2018grounding} which heavily relies on fine-grained annotations in the form of region-sentence pairs. Fine-grained annotations for video grounding are more complicated and labor-intensive as one may need to annotate a spatio-temporal tube (\ie,~label the spatial region in each frame) in a video which semantically corresponds to one language query.

\begin{figure}
    \small
    \textbf{\textit{A brown and white dog is lying on the grass and then it stands up.}}
    \vfill 
    \centering
    \includegraphics[width=1\linewidth]{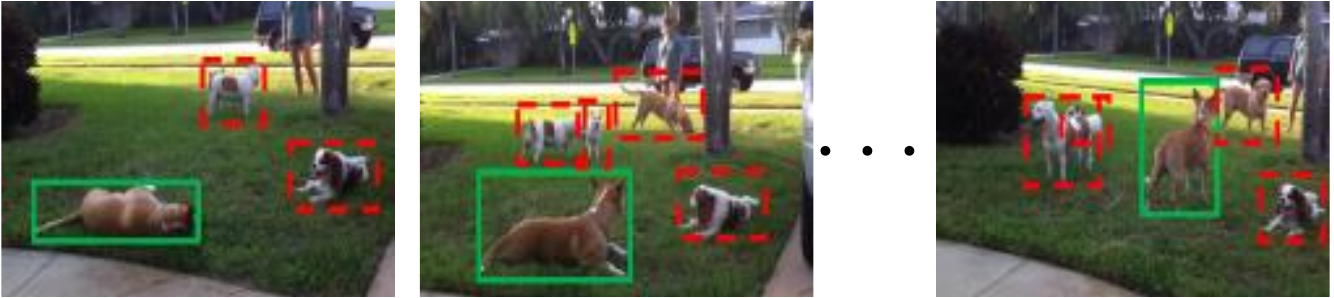}
    \caption{The proposed WSSTG task aims to localize a spatio-temporal tube  (\ie,~the sequence of green bounding boxes) in the video which semantically corresponds to the given sentence, with no reliance on any spatio-temporal annotations during training.
    }
    \label{fig:task}
\end{figure}

To avoid the intensive labor involved in dense annotations,~\cite{huang2018finding} and \cite{zhou2018weakly} considered the problem of weakly-supervised video grounding where only aligned video-sentence pairs are provided without any fine-grained regional annotations. However, they both ground only a noun or pronoun in a static frame of the video. As illustrated in Fig.~\ref{fig:task}, it is difficult to distinguish the target dog (denoted by the green box) from other dogs (denoted by the red boxes) if we attempt to ground only the noun ``\texttt{\small dog}'' in one single frame of the video. The main reason is that the textual description of ``\texttt{\small dog}'' is not sufficiently expressive and the visual appearance in one single frame cannot characterize the spatio-temporal dynamics~(\eg,~the action and movements of the ``\texttt{\small dog}'').  


In this paper, we introduce a novel task, referred to as \emph{weakly-supervised spatio-temporally grounding sentence in video  (WSSTG)}. Specifically, given a natural sentence and a video, we aim to localize a spatio-temporal tube (\ie,~a sequence of bounding boxes), referred to as an instance, in the video which semantically matches the given sentence (see Fig.~\ref{fig:task}). During training, we do not rely on any fine-grained regional annotations. Compared with existing weakly-supervised video grounding problems~\cite{zhou2018weakly, huang2018finding}, our proposed WSSTG task has the following two advantages and challenges. First, we aim to ground a natural sentence instead of just a noun or pronoun, which is more comprehensive and flexible. As illustrated in Fig.~\ref{fig:task}, with a detailed description like ``\texttt{\small{lying on the grass and then it stands up}}'', the target dog (denoted by green boxes) can be localized without ambiguity. However, how to comprehensively capture the semantic meaning of a sentence and ground it in a video, especially in a weakly-supervised manner, poses a challenge. Second, compared with one bounding box in a static frame, a spatio-temporal tube (denoted by a sequence of green bounding boxes in Fig.~\ref{fig:task}) presents the temporal movements of ``\texttt{\small{dog}}'', which can characterize its visual dynamics and thereby semantically match the given sentence. However, how to exploit and model the spatio-temporal characteristics of the tubes as well as their complicated relationships with the sentence poses another challenge.

To handle the above challenges, we propose a novel model realized within the multiple instance learning framework~\cite{karpathy2015deep,Tang_2017_CVPR,tang2018pcl}. First, a set of instance proposals are extracted from a given video. Features of the instance proposals and the sentence are then encoded by a novel attentive interactor that exploits their fine-grained relationships to generate semantic matching behaviors. Finally, we propose a diversity loss, together with a ranking loss, to train the whole model. During testing, the instance proposal which exhibits the strongest semantic matching behavior with the given sentence is selected as the grounding result.

To facilitate our proposed WSSTG task, we contribute a new grounding dataset, called {VID-sentence}, by providing sentence descriptions for the instances of the ImageNet video object detection dataset (VID)~\cite{russakovsky2015imagenet}. Specifically, $7,654$ instances of $30$ categories from $4,381$ videos in VID are extracted. For each instance, annotators are asked to provide a natural sentence describing its content. Please refer to Sec.~\ref{sec:label} for more details about the dataset.

Our main contributions can be summarized as follows. 
1) We tackle a novel task, namely weakly-supervised spatio-temporally video grounding (WSSTG), which localizes a spatio-temporal tube in a given video that semantically corresponds to a given natural sentence, in a weakly-supervised manner.
2) We propose a novel attentive interactor to exploit fine-grained relationships between instances and the sentence to characterize their matching behaviors. A diversity loss is proposed to strengthen the matching behaviors between reliable instance-sentence pairs and penalize the unreliable ones during training. 
3) We contribute a new dataset, named as VID-sentence, to serve as a benchmark for the novel WSSTG task. 
4) Extensive experimental results are analyzed, which illustrate the superiority of our proposed method.

\section{Related Work}
\textbf{Grounding in Images/Videos.} Grounding in images has been popular in the research community over the past decade~\cite{kong2014you,matuszek2012joint,hu2016natural,wang2016learning,wang2016structured,li2017deep,cirik2018using,sadeghi2011recognition,zhang2017discriminative,xiao2017weakly,chen2019localizing,chen2018temporally}. In recent years, researchers also explore grounding in videos. \newcite{yu2015sentence} grounded objects in constrained videos by leveraging weak semantic constraints implied by a sequence of sentences.  \newcite{vasudevan2018object} grounded objects in the last frame of stereo videos with the help of text, motion cues, human gazes and spatial-temporal context. However, fully supervised grounding requires intensive labor for regional annotations, especially in the case of videos. 

\textbf{Weakly-Supervised Grounding.}
To avoid the intensive labor involved in regional annotations, weakly-supervised grounding has been proposed where only image-sentence or video-sentence pairs are needed. It was first studied in the image domain~\cite{zhao2018weakly,rohrbach2016grounding}. Later, given a sequence of transcriptions and their corresponding video clips as well as their temporal alignment, \newcite{huang2018finding} grounded nouns/pronouns in specific frames by constructing a visual grounded action graph. The work closest to ours is~\cite{zhou2018weakly}, in which the authors grounded a noun in a specific frame by considering object interactions and loss weighting given one video and one text input. In this work, we also focus on grounding in a video-text pair. However, different from~\cite{zhou2018weakly} whose text input consists of nouns/pronouns and output is a bounding box in a specific frame, we aim to ground a natural sentence and output a spatio-temporal tube in the video.


~\begin{figure}
    \centering
    \includegraphics[width=\linewidth]{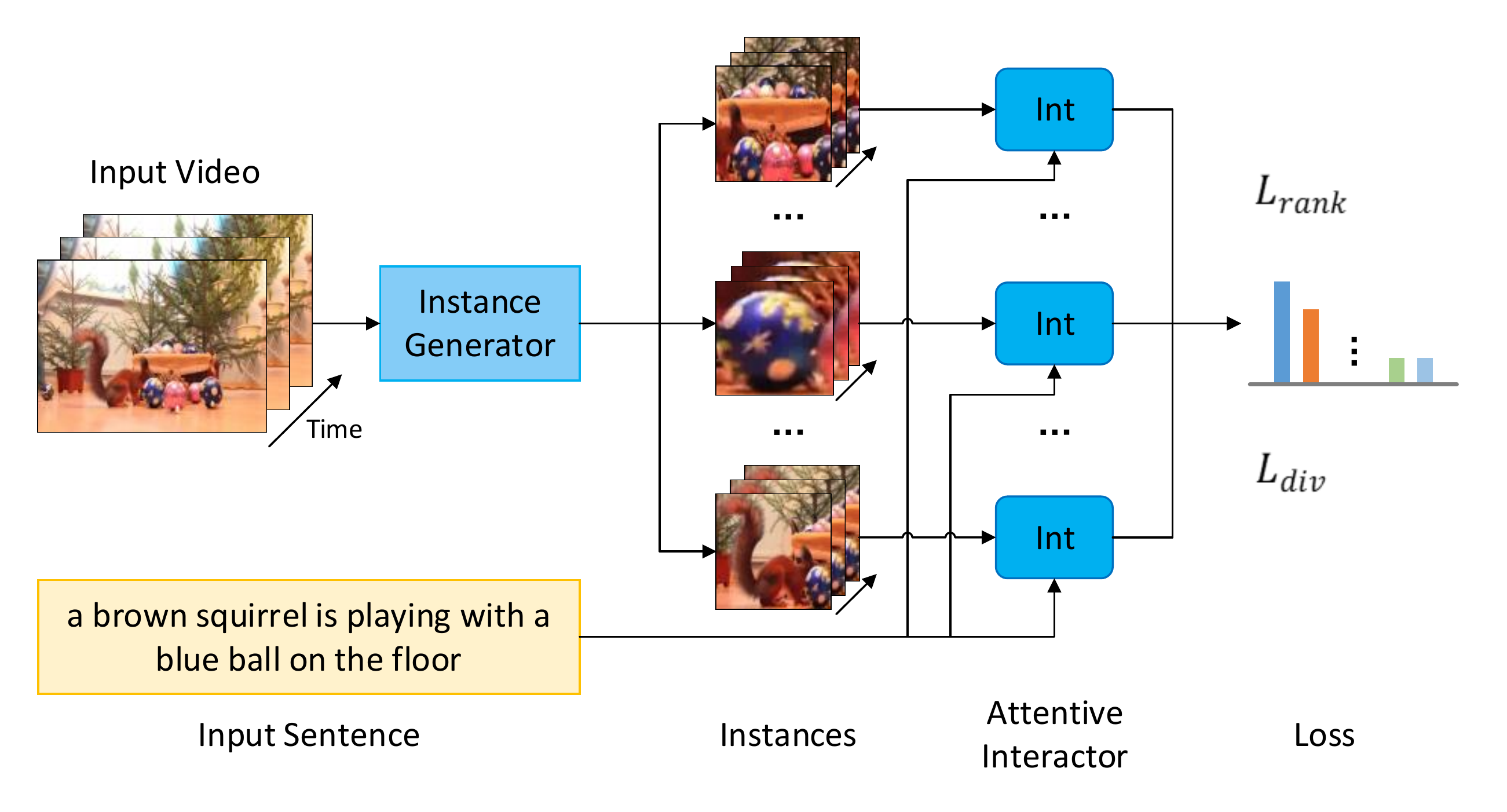}
    \caption{The architecture of our model. An instance generator is used to produce spatio-temporal instances. An attentive interactor is proposed to exploit the complicated relationships between instances and the sentence. Multiple instance learning is used to train the model with a ranking loss and a diversity loss.}
    \label{fig:frm}
\end{figure}

\section{Method}
Given a natural sentence query ${q}$ and a video ${v}$, our proposed WSSTG task aims to localize a spatio-temporal tube, referred to as an instance, $p=\{b_t \}_{t=1}^T$ in the video sequence, where $b_t$ represents a bounding box in the $t$-th frame and $T$ denotes the total number of frames. The localized instance should semantically correspond to the sentence query ${q}$. As WSSTG is carried out in a weakly-supervised manner, only aligned video-sentence pairs $\{v, q\}$ are available with  no fine-grained regional annotations during training. In this paper, we cast the WSSTG task as a multiple instance learning problem~\cite{karpathy2015deep}. Given a video $v$, we first generate a set of instance proposals by an instance generator~\cite{gkioxari2015finding}. We then identify which instance semantically matches the natural sentence query $q$.   

We propose a novel model for handling the WSSTG task. It consists of two components, namely an instance generator and an attentive interactor (see Fig.~\ref{fig:frm}). The instance  generator links bounding boxes detected in each frame into instance proposals (see Sec.~\ref{sec:3.1}). The attentive interactor exploits the complicated relationships between instance proposals and the given sentence to yield their matching scores (see Sec.~\ref{sec:3.2}). The proposed model is optimized with a ranking loss $\mathcal{L}_{rank}$ and a novel diversity loss $\mathcal{L}_{div}$ (see Sec.~\ref{sec:3.3}). Specifically, $\mathcal{L}_{rank}$ aims to distinguish aligned video-sentence pairs from the unaligned ones, while $\mathcal{L}_{div}$ targets strengthening the matching behaviors between reliable instance-sentence pairs and penalizing the unreliable ones from the aligned video-sentence pairs.


\subsection{Instance Extraction}
\label{sec:3.1}
\noindent\textbf{Instance Generation.} As shown in Fig.~\ref{fig:frm}, the first step of our method is to generate instance proposals. Similar to \cite{zhou2018weakly}, the region proposal network from Faster-RCNN~\cite{ren2015faster} is used to detect frame-level bounding boxes with corresponding confidence scores, which are then linked to produce spatio-temporal tubes.

Let $b_t$ denote a detected bounding box at time $t$ and $b_{t+1}$ denote another box at time \mbox{$t+1$}. Following~\cite{gkioxari2015finding}, we define the linking score $s_{l}$ between $b_t$ and $b_{t+1}$ as
\begin{equation}
    \small
    s_l(b_t, b_{t+1}) = s_{c}(b_t) + s_{c}(b_{t+1}) + \lambda\cdot \text{IoU}(b_t, b_{t+1}),
\end{equation}
where $s_{c}(b)$ is the confidence score of $b$, $\text{IoU}(b_t, b_{t+1})$ is the intersection-over-union (IoU) of $b_t$ and $b_{t+1}$, and $\lambda$ is a balancing scalar which is set to 0.2 in our implementation.

As such, one instance proposal $p^n$ can be viewed as a path $\{b^n_t\}_{t=1}^T$ over the whole video sequence with energy $E(p^n)$ given by

\begin{equation}
    \small
    E(p^n) = \frac{1}{T-1}\sum_{t=1}^{T-1}s_l(b^n_t, b^n_{t+1}). 
\end{equation}
We identify the instance proposal with the maximal energy by the Viterbi algorithm~\cite{gkioxari2015finding}. We keep the identified instance proposal and remove all the bounding boxes associated with it. We then repeat the above process until there is no bounding box left. This results in a set of instance proposals $P = \{ p^n\}_{n=1}^N$, with $N$ being the total number of proposals.

\textbf{Feature Representation.} Since an instance proposal consists of bounding boxes in consecutive video frames, we use I3D~\cite{carreira2017quo} and Faster-RCNN to generate the RGB sequence feature \emph{I3D-RGB}, the flow sequence feature \emph{I3D-Flow}, and the frame-level \emph{RoI pooled feature}, respectively. Note that it is not effective to encode each bounding box as an instance proposal may include thousands of bounding boxes. We therefore evenly divide each instance proposal into $t_p$ segments and average the features within each segment. $t_p$ is set to 20 for all our experiments. We concatenate all three kinds of visual features before feeding it into the following attentive interactor. Taking each segment as a time step, each proposal $p$ is thereby represented as $\mathbf{F}_p\in \mathbb{R}^{t_p\times d_p}$, a sequence of $d_p$ dimensional concatenated visual features at each step.

~\begin{figure}
    \centering
      \includegraphics[width=1\linewidth]{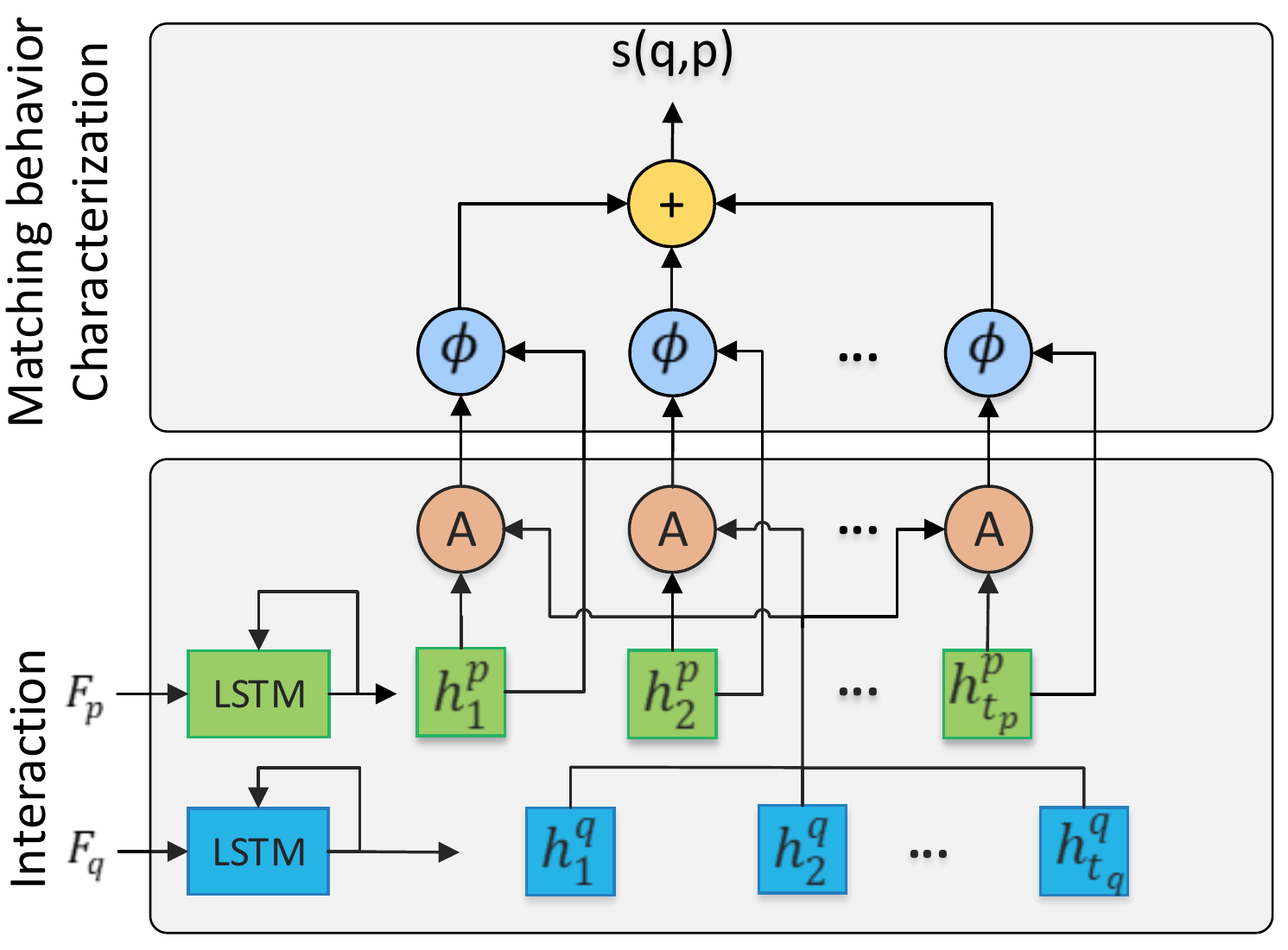}
    \caption{The architecture of the attentive interactor. It consists of two components, namely interaction and matching behavior characterization. \textcircled{\small{A}} denotes the attention mechanism in Eqs.~(\ref{eq:att1}-\ref{eq:att3}). \textcircled{$\phi$} denotes the function in Eq.~(\ref{eq:att3_1}).
    \label{fig:int}}
\end{figure}

\subsection{Attentive Interactor}
\label{sec:3.2}
With the instance proposals from the video and the given sentence query, we propose a novel attentive interactor to characterize the matching behaviors between each proposal and the sentence query. Our attentive interactor consists of two coupling components, namely interaction and matching behavior characterization (see Fig.~\ref{fig:int}). 

Before diving into the details of the interactor, we first introduce the representation of the query sentence $q$. We represent each word in $q$ using the $300$-dimensional word2vec~\cite{mikolov2013distributed} and omit words that are not in the dictionary. In this way, each sentence $q$ is represented as $\mathbf{F}_{q} \in \mathbb{R}^{t_q\times d_q}$, where $t_q$ is the total number of words in the sentence and $d_q$ denotes the dimension of the word embedding.

\subsubsection{Interaction}
Given the sequential visual features $\mathbf{F}_{p}\in \mathbb{R}^{t_p\times d_p}$ of one candidate instance and the sequential textual features $\mathbf{F}_{q}\in \mathbb{R}^{t_q\times d_q}$ of the query sentence, we propose an interaction module to exploit their complicated matching behaviors in a fine-grained manner.
First, two long short-term memory networks (LSTMs)~\cite{hochreiter1997long} are utilized to encode the instance proposal and sentence, respectively:
 \begin{equation}
    \small
     \begin{aligned}
         \mathbf{h}_t^p&=\text{LSTM}_p(\mathbf{f}_t^p,\mathbf{h}_{t-1}^p), \\
         \mathbf{h}_t^q&=\text{LSTM}_q(\mathbf{f}_t^q,\mathbf{h}_{t-1}^q),
     \end{aligned}
 \end{equation}
where $\mathbf{f}_t^p$ and $\mathbf{f}_t^q$ are the $t$-th row representations in $\mathbf{F}_p$ and $\mathbf{F}_q$, respectively. Due to the natural characteristics of LSTM, $\mathbf{h}_{t}^p$ and  $\mathbf{h}_{t}^q$, as the yielded hidden states, encode and aggregate the contextual information from the sequential representation, and thereby yield more meaningful and informative visual features $\mathbf{H}_p=\{\mathbf{h}^p_{t}\}_{t=1}^{t_p}$ and sentence representations $\mathbf{H}_q=\{\mathbf{h}^{q}_t\}_{t=1}^{t_q}$.

Different from~\cite{rohrbach2016grounding,zhao2018weakly} which used only the last hidden state $\mathbf{h}^q_{t_q}$ as the feature embedding for the query sentence, we generate visually guided sentence features $\mathbf{H}_{qp}=\{\mathbf{h}^{qp}_{t}\}_{t=1}^{t_p}$ by exploiting their fine-grained relationships based on $\mathbf{H}_q$ and $\mathbf{H}_p$. Specifically, given the $i$-th visual feature $\mathbf{h}_i^p$, an attention mechanism~\cite{xu2015show} is used to adaptively summarize $\mathbf{H}_q=\{\mathbf{h}_t^q\}_{t=1}^{t_q}$ with respect to $\mathbf{h}_i^p$:

{
    \small
    \begin{align}
    e_{i,j} &=\mathbf{w^\mathrm{T}}\tanh{(\mathbf{W}^q\mathbf{h}_j^q+\mathbf{W}^p\mathbf{h}_i^p+\mathbf{b}_1)}+b_2, \label{eq:att1} \\ 
    a_{i, j} &= \frac{\exp(e_{i,j})}{\sum_{j'=1}^{t_q}\exp(e_{i,j'})}, \label{eq:att2} \\
    \mathbf{h}_{i}^{qp} &= \sum_{j=1}^{t_q}a_{i,j}\mathbf{h}_j^q, \label{eq:att3}
    \end{align}
    
    }
where $\mathbf{W}^q \in \mathbb{R}^{K\times D_q}$, $\mathbf{W}^p \in \mathbb{R}^{K\times D_p}$, $\mathbf{b}_1 \in \mathbb{R}^{K}$ are the learnable parameters that map visual and sentence features to the same $K$-dimension space. $\mathbf{w} \in \mathbb{R}^{K}$ and $b_2 \in \mathbb{R}$ work on the coupled textual and visual features and yield their affinity scores. With respect to $\mathbf{W}^p\mathbf{h}_i^p$ in Eq.~(\ref{eq:att1}), the generated visually guided sentence feature $\mathbf{h}_i^{qp}$ pays more attention on the words more correlated with $\mathbf{h}_i^p$ by adaptively summarizing $\mathbf{H}_q=\{\mathbf{h}_t^q\}_{t=1}^{t_q}$. 

Owning to the attention mechanism in Eqs. (\ref{eq:att1}-\ref{eq:att3}), our proposed interaction module makes each visual feature interact with all the sentence features and attentively summarize them together. As such, fine-grained relationships between the visual and sentence representations are exploited. 
 
\subsubsection{Matching Behavior Characterization}
After obtaining a set of visually guided sentence features $\mathbf{H}_{qp}=\{\mathbf{h}^{qp}_{t}\}_{t=1}^{t_p}$, we characterize the fine-grained matching behaviors between the visual and sentence features. Specifically, the matching behavior between the $i$-th visual and sentence features is defined as
 \begin{equation}
    \small
     s_i(\mathbf{h}_i^p, \mathbf{h}_i^{qp}) = \phi(\mathbf{h}_i^p,\mathbf{h}_i^{qp}).
     \label{eq:att3_1}
 \end{equation}
 The instantiation of $\phi$ can be realized by different approaches,  such as multi-layer perceptron (MLP), inner-product, or cosine similarity. In this paper, we use cosine similarity between $\mathbf{h}_i^p$ and $\mathbf{h}_i^{qp}$ for simplicity. 
 Finally, we define the matching behavior between an instance proposal $p$ and the sentence $q$ 
 as

 \begin{equation}
    \small
     s(q, p) = \frac{1}{t_p}\sum_{i=1}^{t_p}s_i(\mathbf{h}_i^p, \mathbf{h}_i^{qp}).
     \label{eq:att4}
 \end{equation}

\subsection{Training}
\label{sec:3.3}
For the WSSTG task,
since no regional annotations are available during the training, we cannot optimize the framework in a fully supervised manner. We, therefore, resort to MIL to optimize the proposed network based on the obtained matching behaviors of the instance-sentence pairs.
Specifically, our objective function is defined as
\begin{equation}
    \small
    \mathcal{L} = \mathcal{L}_{rank} + \beta~\mathcal{L}_{div},
    \label{eq:loss}
\end{equation}
where $\mathcal{L}_{rank}$ is a ranking loss, aiming at distinguishing aligned video-sentence pairs from the unaligned ones. $\mathcal{L}_{div}$ is a novel diversity loss, which is proposed to strengthen the matching behaviors between reliable instance-sentence pairs and penalize the unreliable ones from the aligned video-sentence pair. $\beta$ is a scalar which is set to 1 in all our experiments.

\noindent\textbf{Ranking Loss.} Assume that $\{v, q\}$ is a semantically aligned video-sentence pair. We define the visual-semantic matching score $S$ between $v$ and $q$ as
\begin{equation}
    \small
    S(v, q) = \max~s(q, p_n)~,n=1,...,N\, ,
\end{equation}
where $p_n$ is the $n$-th proposal generated from the video $v$, $s(q, p_n)$ is the matching behavior computed by Eq.~(\ref{eq:att4}), and $N$ is the total number of instance proposals.

Suppose that $v'$ and $q'$ are negative samples that are not semantically correlated with 
$q$ and $v$, respectively. Inspired by~\cite{karpathy2015deep}, we define the ranking loss as
\begin{equation}
\small 
\begin{split}
    \mathcal{L}_{rank} &= \sum_{v \neq v'}\sum_{q \neq q'}[\max(0,~S(v, q')-S(v, q)+ \Delta)+\\
    & \max(0,~S(v', q)-S(v, q)+ \Delta)],
\end{split}
\label{loss:rank}
\end{equation}
where $\Delta$ is a margin which is set to 1 in all our experiments. $\mathcal{L}_{rank}$ directly encourages the matching scores of aligned video-sentence pairs to be larger than those of unaligned pairs. 

\noindent\textbf{Diversity Loss.} One limitation of the ranking loss defined in Eq.~(\ref{loss:rank}) is that it does not consider the matching behaviors between the sentence and different instance proposals extracted from an aligned video. A prior for video grounding is that only a few instance proposals in the paired video are semantically aligned to the query sentence, while most of the other instance proposals are not. Thus, it is desirable to have a diverse distribution of the matching behaviors $\{s(q, p_n)\}_{n=1}^N$.

To encourage a diverse distribution of $\{s(q, p_n)\}_{n=1}^N$, we propose a diversity loss $\mathcal{L}_{div}$ to strengthen the matching behaviors between reliable instance-sentence pairs and penalize the unreliable ones during training. Specifically, we first normalize $\{s(q, p_n)\}_{n=1}^N$ 
by softmax
\begin{equation}
\small
    s'(q, p_n) = \frac{\exp(s(q, p_n))}{\sum_{n'=1}^{N} \exp(s(q, p_{n'}))}, \label{eq:div1}
\end{equation}
and then penalize the entropy of the distribution of $\{s'(q, p_n)\}_{n=1}^N$ by defining the diversity loss as
\begin{equation}
    \small
    \mathcal{L}_{div} = -\sum_{n=1}^{N}{s'(q, p_n)log(s'(q, p_n))}. \label{eq:div2}
\end{equation}
Note that the smaller $\mathcal{L}_{div}$ is, the more diverse $\{s(q, p_n)\}_{n=1}^N$ will be, which implicitly encourages the matching scores of semantically aligned instance-sentence pairs being larger than those of the misaligned pairs.

\subsection{Inference}
Given a testing video and a query sentence, we extract candidate instance proposals, and characterize the matching behavior between each instance proposal and the sentence by the proposed attentive interactor. The instance with the strongest matching behavior is deemed the result of the WSSTG task.

\section{VID-sentence Dataset}
\label{sec:label}
A main challenge for the WSSTG task is the lack of suitable datasets. Existing datasets like TACoS~\cite{regneri2013grounding} and YouCook~\cite{DaXuDoCVPR2013} are unsuitable as they do not provide spatio-temporal annotations for target instances in the videos, which are necessary for the WSSTG task for evaluation. To the best of our knowledge, the most suitable existing dataset is the Person-sentence dataset provided by~\cite{yamaguchi2017spatio}, which is used for spatio-temporal person search among videos. However, this dataset is too simple for the WSSTG task since it contains only people in the videos. 
To this end, we contribute a new dataset by annotating videos in ImageNet video object detection dataset (VID)~\cite{russakovsky2015imagenet} with sentence descriptions. We choose VID as the visual materials for two primary reasons. First, it is one of the largest video detection datasets containing videos of diverse categories in complicated scenarios. Second, it provides dense bounding-box annotations and instance IDs which help avoid labor-intensive annotations for spatio-temporal regions of the validation/testing set.

\begin{figure}[t]
    \small
    \textbf{\textit{A red bus is making a turn on the road}}
    \vfill
    \includegraphics[width =1\linewidth,height=0.16\linewidth]{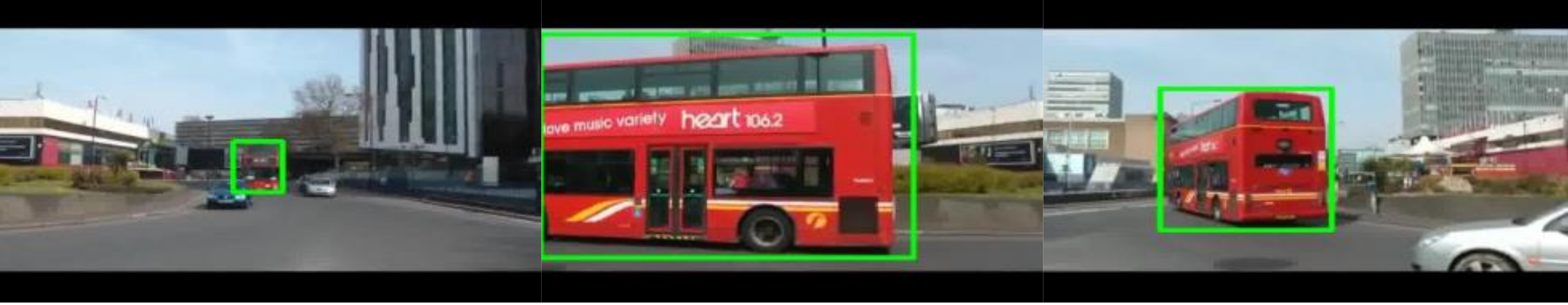}
    \textbf{\textit{A large elephant runs in the water from left to right}}
    \vfill
    \includegraphics[width =1\linewidth,height=0.16\linewidth]{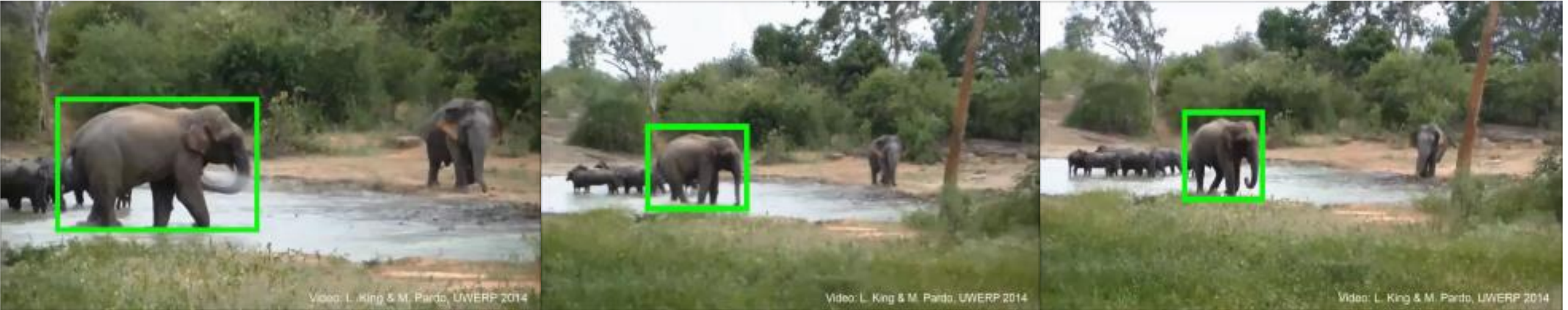}
 \caption{Samples of the newly constructed VID-sentence dataset. Sentences are shown on the top of images and the associated target instances are enclosed with green bounding boxes.}
    \label{fig:samples}
\end{figure}
\noindent\textbf{VID-sentence Annotation.}
With $30$ categories, VID contains $3826$, $555$ and $937$ videos for training, validation and testing respectively. We first divide videos in training and validation sets\footnote{Testing set is omitted as its spatial-temporal annotations are unavailable} into trimmed videos based on the provided instance IDs, and delete videos less than $9$ frames. As such, there remain $9,029$ trimmed videos in total. In each trimmed video, one instance is identified as a sequence of bounding boxes. A group of annotators are asked to provide sentence descriptions for the target instances. Each target instance is annotated with one sentence description. An instance is discarded if it is too difficult to provide a unique and precise description.
After annotation, there are $7,654$ videos with sentence descriptions. We randomly select $6,582$ videos as the training set, and evenly split the remaining videos into the validation and testing sets (\ie, each contains $536$ videos). Some examples from the VID-sentence dataset are shown in Fig.~\ref{fig:samples}. 


\noindent\textbf{Dataset Statistics.}
To summarize, the created dataset has $6,582/536/536$ spatio-temporal instances with descriptions for training/validation/testing. It covers all $30$ categories in VID, such as ``\texttt{\small car}'', ``\texttt{\small monkey}'' and ``\texttt{\small watercraft}''. The size of the vocabulary is $1,823$ and the average length of the descriptions is $13.2$. 
Table~\ref{tb:dataset} shows the statistics of our constructed VID-sentence dataset. Compared with the Person-sentence dataset, our VID-sentence dataset has a similar description length but includes more instances and categories. 

It is important to note that, although VID provides regional annotations for the training set, these annotations are not used in any of our experiments since we focus on weakly-supervised spatio-temporal video grounding. 

\section{Experiments}
In this section, we first compare our method with different kinds of baseline methods on the created VID-sentence dataset, followed by the ablation study. Finally, we show how well our model generalizes on the Person-sentence dataset.
\subsection{Experimental Settings}
\noindent\textbf{Baseline Models}.
Existing weakly-supervised video grounding methods~\cite{huang2018finding, zhou2018weakly} are not applicable to the WSSTG task. \newcite{huang2018finding} requires temporal alignment between a sequence of transcription descriptions and the video segments to ground a noun/pronoun in a certain frame, while \newcite{zhou2018weakly} mainly grounds nouns/pronouns in specific frames of videos. As such, we develop three baselines based on DVSA~\cite{karpathy2015deep}, GroundeR~\cite{rohrbach2016grounding}, and a variant frame-level method modified from \cite{zhou2018weakly} for performance comparisons. Following recent grounding methods like~\cite{rohrbach2016grounding,chen2018knowledge}, we use the last hidden state of an LSTM encoder as the sentence embedding for all the baselines.

Since DVSA and GroundeR are originally proposed for image grounding, in order to adapt to video, we consider three methods to encode visual features $\mathbf{F}_{p}\in \mathbb{R}^{t_p\times d_p}$ including averaging (Avg), NetVLAD~\cite{arandjelovic2016netvlad}, and LSTM. For the variant baseline modified from \cite{zhou2018weakly}, we densely predict each frame to generate a spatio-temporal prediction.

\begin{table}[t]
\centering
\small
\renewcommand{\arraystretch}{1}
\setlength{\tabcolsep}{0.7em}
\begin{tabular}{|l|ccccc|}
\hline
&\multicolumn{3}{c}{Instance Num.} & Des. & \multirow{2}{*}{Categories}  \\ 
   &  train & val & test &  length &   \\     
 \hline
 \hline
 Person &  5,437  & 313 & 323  &  13.1 & 1\\ 
 Ours   &  6,582  & 536 & 536  &  13.2 & 30 \\ \hline
\end{tabular}
\caption{Statistics of the VID-sentence dataset and previous Person-sentence dataset~\newcite{yamaguchi2017spatio}.}
\label{tb:dataset}
\end{table}

\noindent\textbf{Implementation Details}. Similar to \cite{zhou2018weakly}, we use the region proposal network from Faster-RCNN pretrained on MSCOCO~\cite{lin2014microsoft} to extract frame-level region proposals. For each video, we extract $30$ bounding boxes for each frame and link them into $30$ spatio-temporal tubes with the method~\cite{gkioxari2015finding}. 
We map the word embedding to $512$-dimension before feeding it to the LSTM encoder. Dimension of the hidden state of all the LSTMs is set to $512$. Batch size is $16$, \ie, $16$ videos with total $480$ instance proposals and $16$ corresponding sentences. We construct positive and negative video-sentence pairs for training within a batch for efficiency, \ie, roughly 16 positive pairs and 240 negative pairs for the triplet construction. SGD is used to optimize the models with a learning rate of 0.001 and momentum of 0.9. We train all the models with 30 epochs. Please refer to supplementary materials for more details.

\noindent\textbf{Evaluation Metric}. We use the bounding box localization accuracy for evaluation. An output instance is considered as ``accurate" if the overlap between the detected instance and the ground-truth is greater than a threshold $\eta$. The definition of the overlap is the same as \cite{yamaguchi2017spatio}, \ie, the average overlap of the bounding boxes in annotated frames. $\eta$ is set to $0.4$, $0.5$, $0.6$ for extensive evaluations.


\begin{figure*}[t]
\small
Description: \textbf{\textit{The white car is running from left to right on the left side of the road.}}
\vfill
\begin{minipage}[t]{0.5\linewidth}
\centering
\includegraphics[width =0.9\textwidth,height=0.16\textwidth]{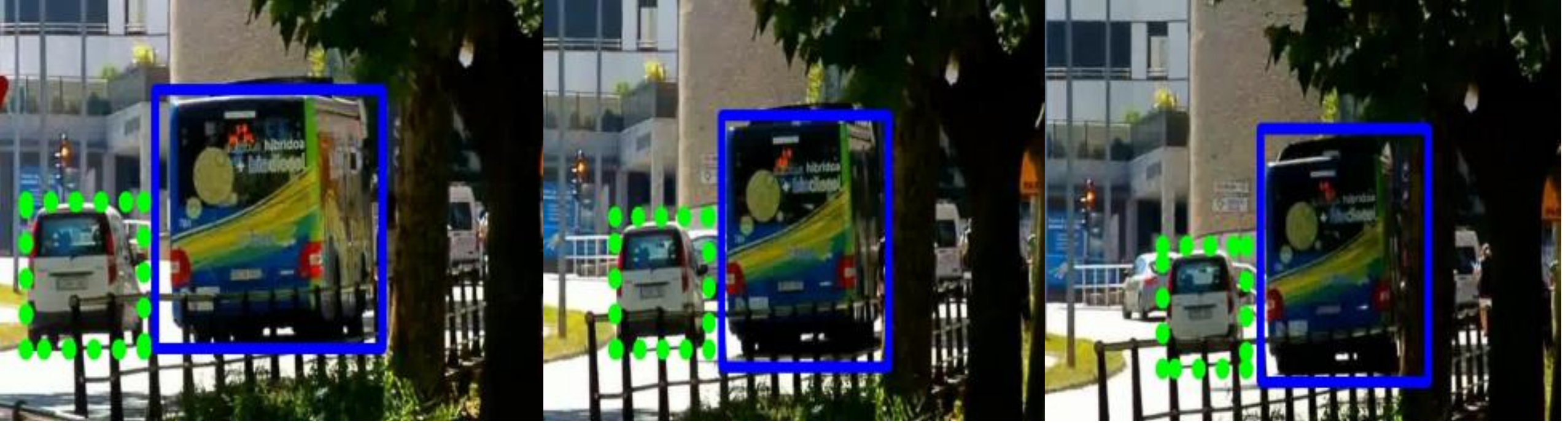}
\small{\centering{DVSA+LSTM, IoU: 0.172}}
\end{minipage}
\begin{minipage}[t]{0.5\linewidth}
\centering
\includegraphics[width =0.9\textwidth,height=0.16\textwidth]{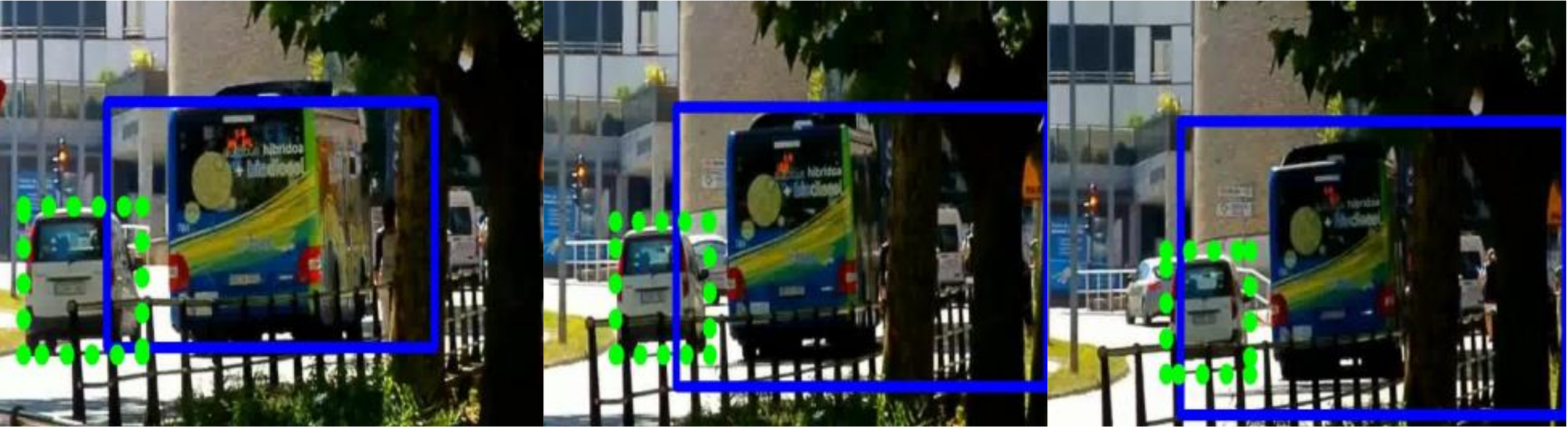}
\small{\centering{GroundeR+LSTM, IoU: 0.042}}
\end{minipage}
\vfill
\begin{minipage}[t]{0.5\linewidth}
\centering
\includegraphics[width =0.9\textwidth,height=0.16\textwidth]{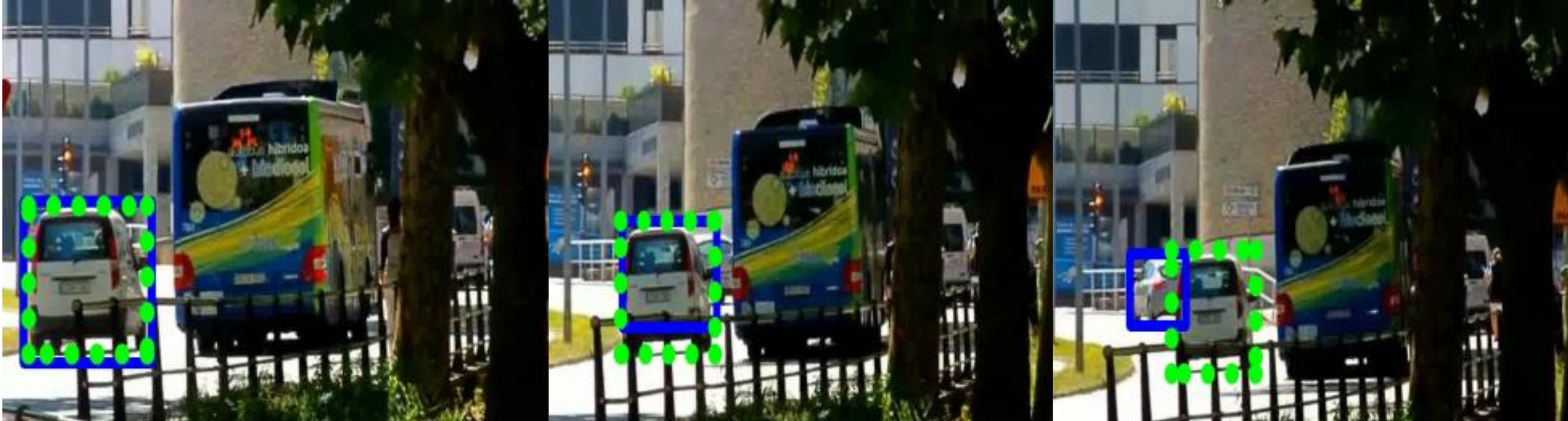}
\\
\small{\centering{\newcite{zhou2018weakly}, IoU: 0.413}}
\end{minipage}
\begin{minipage}[t]{0.5\linewidth}
\centering
\includegraphics[width =0.9\textwidth,height=0.16\textwidth]{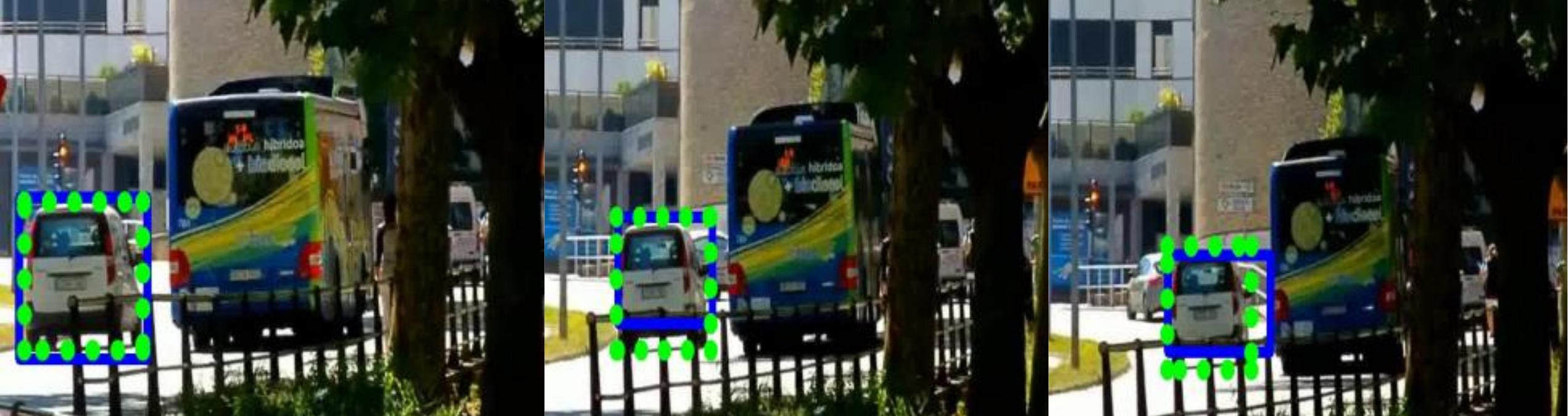}\\
\small{\centering{Ours, IoU: 0.604}}
\end{minipage}
\caption{An exemplar of the results by different methods. The sentence is shown on the top. Three frames of the detected results and the ground-truth are respectively bounded with blue lines and green dotted lines. IoU scores between the detected instances and the ground-truth are shown below the images. Best viewed on screen.\label{fig:exp4}}
\end{figure*}  
\begin{table}[t]
\centering
\renewcommand{\arraystretch}{1}
\setlength{\tabcolsep}{0.5em}
\small
\begin{tabular}{|l|cccc|}
\hline
\multirow{2}{*}{Methods} & \multicolumn{4}{c|}{Accuracy} \\ 
                         & 0.4  & 0.5  & 0.6  & Average \\ 
\hline
\hline
Random                   & 8.0  & 4.3  & 2.1 & 4.8     \\ 
Proposal upper bound      & 58.6 & 47.2 & 36.9 & 47.6   \\ 
\hline
\hline
DVSA+Avg                                         & 36.2 & 29.7 & 23.5 & 29.8    \\
DVSA+NetVLAD             & 31.2 & 24.8 & 18.5 & 24.8    \\
DVSA+LSTM                    & 38.2 & 31.2 & 23.5 & 31.0    \\ 
GroundeR+Avg                               & 36.7 & 31.9 & 25.0 & 31.2    \\
GroundeR+NetVLAD    & 26.1 & 22.2 & 15.1 & 21.1    \\
GroundeR+LSTM            & 36.8 & 31.2 & 24.1 & 30.7    \\ 
\newcite{zhou2018weakly}                                         &   41.6   &   33.8   & 27.1 &34.2 \\  
\hline
\hline
Ours                     & \textbf{44.6} & \textbf{38.2} & \textbf{28.9} & \textbf{37.2} 
\\ \hline
\end{tabular}
\caption{Performance comparisons on the proposed VID-sentence dataset. The top entry of all the methods except the upper bound is highlighted in boldface.}
\label{exp:vid}
\end{table}

\subsection{Performance Comparisons}
Table~\ref{exp:vid} shows the performance comparisons between our model and the baselines. We additionally show the performance of randomly choosing an instance proposal and the upper bound performance of choosing the instance proposal of the largest overlap with the ground-truth.

The results suggest that, 1) models with NetVLAD ~\cite{arandjelovic2016netvlad} perform the worst. We suspect that models based on NetVLAD are complicated and the supervisions are too weak to optimize the models sufficiently well. 2) Models with LSTM embedding achieve only comparable performances compared with models based on simple averagingf. It is mainly due to the fact  that the power of LSTM has not been fully exploited. 3) The variant method of \cite{zhou2018weakly} performs better than both DVSA and GroundeR with various kinds of visual encoding techniques, indicating its power for the task. 4) Our model achieves the best results, demonstrating its effectiveness, showing that our model is better at characterizing the matching behaviors between the query sentence and the visual instances in the video.

To compare the methods qualitatively, we show an exemplar sample in Fig.~\ref{fig:exp4}. Compared with GroundeR+LSTM and DVSA+LSTM, our method identifies a more accurate instance from the candidate instance proposals. Moreover, the instances generated by our method are more temporally consistent compared with the modified frame-level method~\cite{zhou2018weakly}. This can be attributed to the exploitation of the temporal information during instance generation and attentive interactor in our model. 

\begin{table}[t]
\centering
\small
\renewcommand{\arraystretch}{1}
\setlength{\tabcolsep}{0.9em}
\begin{tabular}{|l|cccc|}
\hline
\multirow{2}{*}{Methods} & \multicolumn{4}{c|}{Accuracy}                      \\  
                         & 0.4  & 0.5  & 0.6  & Average \\ 
\hline
\hline
Base                     & 38.2 & 31.2 & 23.5 & 31.0    \\
Base + Div               & 38.4 & 32.5 & 25.0 & 32.0    \\
Base + Int               & 42.4 & 35.1 & 26.1 & 34.5    \\
Full method              & \textbf{44.6} & \textbf{38.2} & \textbf{28.9} & \textbf{37.2} \\ \hline
\end{tabular}
\caption{Ablation study of the proposed attentive interactor and diversity loss.} 
\label{exp:abs}
\end{table}

\subsection{Ablation Study}
To verify the contributions of the proposed attentive interactor and diversity loss, we perform the following ablation study. To be specific, we compare the full method with three variants, including: 1) removing both the attentive interactor and diversity loss, which is equivalent to the DVSA model using LSTM for encoding both the visual features and sentence features, termed as \texttt{Base}; 2) \texttt{Base+Div}, which is formed by introducing the diversity loss; 3) \texttt{Base+Int} with the attentive interactor module.

\begin{figure}[tb]
\centering
\small
\begin{minipage}[t]{0.32\linewidth}
\includegraphics[width =0.9\textwidth,height=0.5\textwidth]{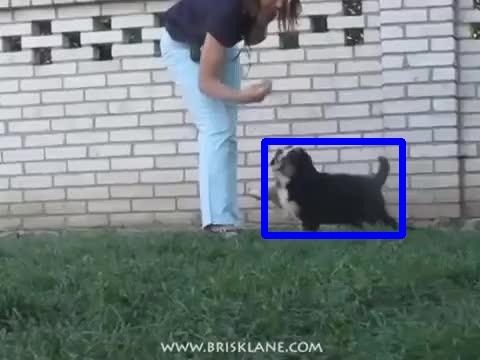}
\centering{segment Id: 0}
\end{minipage}
\begin{minipage}[t]{0.32\linewidth}
\includegraphics[width =0.9\textwidth,height=0.5\textwidth]{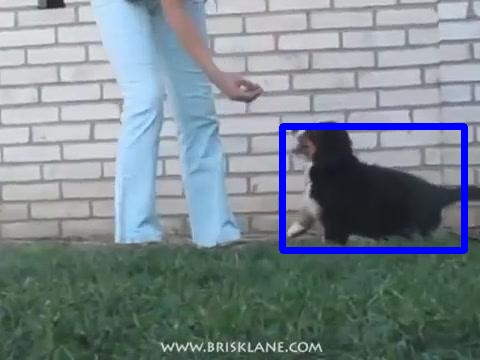}
\centering{segment Id: 1}
\end{minipage}
\begin{minipage}[t]{0.32\linewidth}
\includegraphics[width =0.9\textwidth,height=0.5\textwidth]{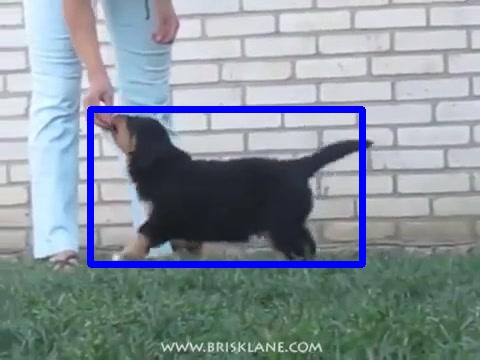}
\centering{segment Id: 2}
\end{minipage}
\hfill
\begin{minipage}[t]{1\linewidth}
\centering
\includegraphics[width=1\linewidth,height=0.25\linewidth]{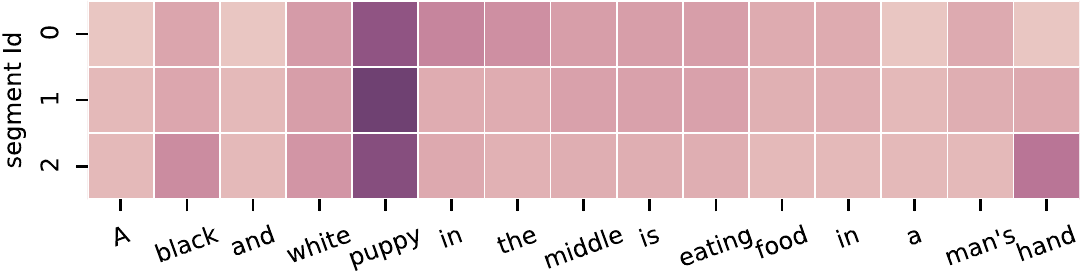}
\end{minipage}
\caption{Visualization of the attentive interaction. On the top, we show an instance highlighted in the blue box in three different segments. On the bottom, we show the corresponding distributions of the attention weights. Darker colors mean larger attentive weights. Intuitively, the attention weight matches well with the visual contents such as ``\texttt{\small{puppy}}'' in all three segments and ``\texttt{\small{hand}}'' in the segment with ID $2$. Best viewed on screen.\label{fig:abs:att}}
\end{figure} 

\begin{figure}[t]
    \centering
    \includegraphics[width=1\linewidth]{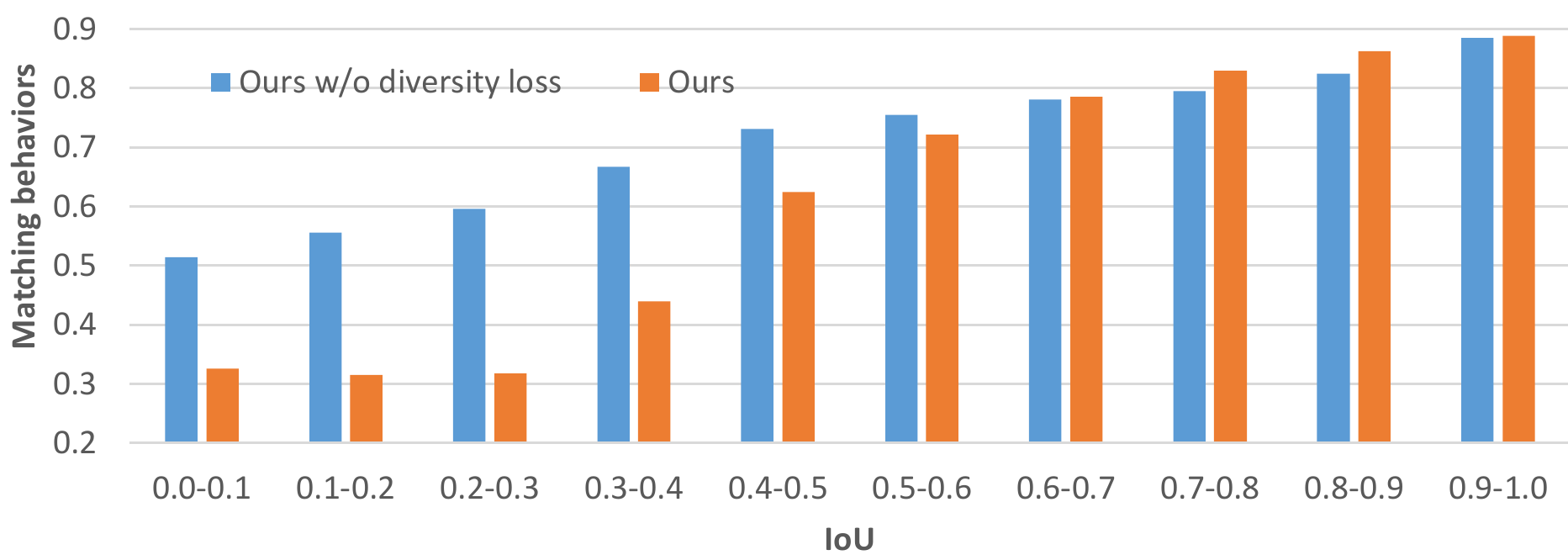}
    \caption{Comparison of the distribution of the matching behaviors of instances.}
    \label{fig:div}
\end{figure}

Table \ref{exp:abs} shows the corresponding results. Compared with \texttt{Base}, both the diversity loss and attentive interactor constantly improve the performance. Moreover, to show the effectiveness of the proposed attentive interactor, we visualize the adaptive weight $a$ in Eq.~\eqref{eq:att2}. As shown in Fig.~\ref{fig:abs:att}, our method adaptively pays more attention to the words that match the instance such as the ``\texttt{\small puppy}" in all three segments and the ``\texttt{\small hand}" in segment with ID 2. To show the effectiveness of the diversity loss, we divide instance proposals in the testing set into $10$ groups based on their IoU scores with the ground-truth and then calculate the average matching behaviors of each group, predicted by counterparts with and without the diversity loss. As shown in Fig.~\ref{fig:div}, the proposed diversity loss $\mathcal{L}_{div}$ penalizes the matching behaviors of the instances of lower IoU 
with ground-truth while strengthens instances of higher IoU.  

\begin{table}[t]
\centering
\renewcommand{\arraystretch}{1}
\setlength{\tabcolsep}{0.5em}
\small
\begin{tabular}{|l|cccc|}
\hline
\multirow{2}{*}{Methods} & \multicolumn{4}{c|}{Accuracy}                      \\  
                         & 0.4  & 0.5  & 0.6  & Average \\ 
\hline
\hline
Random                   & 15.1 & 7.2  & 3.5  &  8.6   \\ 
Proposal upper bound      & 89.8 & 79.9 & 64.1 & 77.9  \\ 
\hline
\hline
DVSA+Avg                                         & 39.8 & 30.3 & 19.7 & 29.9     \\
DVSA+NetVLAD             & 34.1 & 25.0 & 18.3 & 25.8     \\
DVSA+LSTM                     & 42.7 & 30.2 & 20.0 & 31.0     \\ 
GroundeR+Avg                                & 45.5  & 32.2 & 21.7 & 33.1     \\
GroundeR+NetVLAD    & 22.1 & 16.1  & 8.6  & 15.6     \\
GroundeR+LSTM            & 39.9 & 28.2  & 17.7 & 28.6     \\ 
\hline
\hline
Ours w/o $\mathcal{L}_{div}$  & 57.9 & 47.7 & 35.6 & 47.1     \\ 
Ours                     & \textbf{62.5} & \textbf{52.0} & \textbf{38.4} & \textbf{51.0}     \\ \hline
\end{tabular}
\caption{Performance comparisons on the Person-sentence dataset \cite{yamaguchi2017spatio}.}
\label{exp:act}
\end{table}
\subsection{Experiments on Person-sentence Dataset}

We further evaluate our model and the baseline methods on the Person-sentence dataset~\cite{yamaguchi2017spatio}. We ignore the bounding box annotations in the training set and carry out experiments for the proposed WSSTG task. For fair comparisons, all experiments are conducted on the visual feature extractor provided by \cite{carreira2017quo}.

Table~\ref{exp:act} shows the results. Similarly, the proposed attentive interactor model (without the diversity loss) outperforms all the baselines. Moreover, the diversity loss further improves the performance. Note that the improvement of our model on this dataset is more significant than that on the VID-sentence dataset. The reason might be that the upper bound performance of the Person-sentence is much higher than that of the VID-sentence ($77.9$ for Person-sentence \textit{versus} $47.6$ for VID-sentence on average). This also suggests that the created VID-sentence dataset is more challenging and more suitable as a benchmark dataset.

\section{Conclusion}
In this paper, we introduced a new task, namely weakly-supervised spatio-temporally grounding natural sentence in video. It takes a sentence and a video as input and outputs a spatio-temporal tube from the video, which semantically matches the sentence, with no reliance on spatio-temporal annotations during training. 
We handled this task based on the multiple instance learning framework. An attentive interactor and a diversity loss were proposed to learn the complicated relationships between the instance proposals and the sentence. Extensive experiments showed the effectiveness of our model. Moreover, we contributed a new dataset, named as VID-sentence, which can serve as a benchmark for the proposed task.  

\bibliography{acl2019}
\bibliographystyle{acl_natbib}

\clearpage
\section{Supplementary Material}
We provide more descriptions of baseline methods and implementation details in this supplementary material section.

\noindent\textbf{Baseline Details.}
We consider three methods to encode visual instance features $\mathbf{F}_{p}\in \mathbb{R}^{t_p\times d_p}$ including averaging (Avg), NetVLAD~\cite{arandjelovic2016netvlad}, and LSTM. For Avg, we simply average all the $t_p$ segments and forward to two fully connected layers. For NetVLAD, we treat $\mathbf{F}_p$ as $t_p$ independent $d_p$-dimension features and use a fully connected layer to obtain output with the desired dimension. For LSTM, we take $\mathbf{F}_p$ as a sequence features of $t_p$ time steps and use the last hidden state of LSTM as the embedded visual representation.

For models based on DVSA, we evaluate the similarity between the spatio-temporal instance and the query sentence with cosine similarity. For models based on GroundeR, we concatenate the representations from the visual encoder and the sentence encoder as the input for the attention network and reconstruction network. For the variant of \cite{zhou2018weakly}, we densely predict each frame in the video to generate a spatio-temporal instance. This baseline is carefully implemented by modifying the original method~\cite{zhou2018weakly} with two aspects. On one hand, we replace the noun encoder with an LSTM to encode natural sentences, since we focus on grounding with natural sentences. On the other hand, we remove the frame-wise loss weighting term as it degrades the performance on the VID-sentence dataset. Such loss term is proposed to penalize the uncertainty of the existence of objects, which is not necessary as the video in our dataset contains the target instances in all frames.

 The output of Avg and Net-VLAD~\cite{arandjelovic2016netvlad} is also set as 512 by a fully connected layer. The number of centers and the dimension of cluster-center for Net-VLAD are $32$ and $128$, respectively.

\noindent\textbf{Implementation Details.}
 We give more details on how to generate instances from videos and extract the corresponding visual feature for each instance. We use the region proposal network from Faster-RCNN~\cite{ren2015faster} to extract $30$ region proposals for each video frame. The Faster-RCNN model is based on ResNet-101~\cite{he2016deep} pretrained on MSCOCO~\cite{lin2014microsoft}. For the frame-level RoI pooled feature, we use the $2048$-dimensional feature from the last fully connected layer of the same Faster-RCNN model. For the I3D features~\cite{carreira2017quo}, we use the model pretrained on Kinetics to extract the RGB sequence features I3D-RGB and the flow sequence features I3D-Flow. For every $64$ consecutive frames, we extract a set of (eight) $1024$-dimensional I3D-RGB features and (eight) $1024$-dimensional I3D-Flow features by the output of the last average pooling layer and dropping the last temporal pooling operation. We compute optical flow with a TV-L1 algorithm~\cite{zach2007duality}. We crop the region proposals from the RGB images and flow images and then resize them to $224 \times 224$ before feeding to I3D. 

\end{document}